\documentclass[sigconf]{acmart}
\usepackage{multirow}
\usepackage{balance}
\usepackage{bm}
\usepackage{microtype}

\AtBeginDocument{%
  \providecommand\BibTeX{{%
    \normalfont B\kern-0.5em{\scshape i\kern-0.25em b}\kern-0.8em\TeX}}}

\copyrightyear{2022}
\acmYear{2022}
\setcopyright{acmcopyright}
\acmConference[SIGIR '22]{Proceedings of the 45th International ACM SIGIR Conference on Research and Development in Information Retrieval}{July 11--15, 2022}{Madrid, Spain}
\acmBooktitle{Proceedings of the 45th International ACM SIGIR Conference on Research and Development in Information Retrieval (SIGIR '22), July 11--15, 2022, Madrid, Spain}
\acmPrice{15.00}
\acmDOI{10.1145/3477495.3531816}
\acmISBN{978-1-4503-8732-3/22/07}
\begin{document}

\fancyhead{}
\title[Generalizing to the Future: Mitigating Entity Bias in Fake News Detection]{Generalizing to the Future: \\Mitigating Entity Bias in Fake News Detection}


\author{Yongchun Zhu$^{1,2}$, Qiang Sheng$^{1,2}$, Juan Cao$^{1,2,*}$, Shuokai Li$^{1,2}$, Danding Wang$^{1,2}$, Fuzhen Zhuang$^{3,4}$}
\affiliation{%
 \institution{$^1$Key Lab of Intelligent Information Processing of Chinese Academy of Sciences (CAS), Institute of Computing Technology, CAS, Beijing 100190, China\\
 $^2$University of Chinese Academy of Sciences, Beijing 100049, China\\
 $^3$Institute of Artificial Intelligence, Beihang University, Beijing 100191, China\\ $^4$SKLSDE, School of Computer Science, Beihang University, Beijing 100191, China\\
 \{zhuyongchun18s, shengqiang18z, caojuan, lishuokai18z, wangdanding\}@ict.ac.cn, zhuangfuzhen@buaa.edu.cn}\country{}}
\thanks{*Juan Cao is the corresponding author.}

\renewcommand{\shortauthors}{Y. Zhu et al.}

\begin{abstract}
The wide dissemination of fake news is increasingly threatening both individuals and society. Fake news detection aims to train a model on the past news and detect fake news of the future. Though great efforts have been made, existing fake news detection methods overlooked the unintended entity bias in the real-world data, which seriously influences models' generalization ability to future data. For example, 97\% of news pieces in 2010-2017 containing the entity `Donald Trump' are real in our data, but the percentage falls down to merely 33\% in 2018. This would lead the model trained on the former set to hardly generalize to the latter, as it tends to predict news pieces about `Donald Trump' as real for lower training loss. In this paper, we propose an entity debiasing framework (\textbf{ENDEF}) which generalizes fake news detection models to the future data by mitigating entity bias from a cause-effect perspective. Based on the causal graph among entities, news contents, and news veracity, we separately model the contribution of each cause (entities and contents) during training. In the inference stage, we remove the direct effect of the entities to mitigate entity bias. Extensive offline experiments on the English and Chinese datasets demonstrate that the proposed framework can largely improve the performance of base fake news detectors, and online tests verify its superiority in practice. To the best of our knowledge, this is the first work to explicitly improve the generalization ability of fake news detection models to the future data. The code has been released at \url{https://github.com/ICTMCG/ENDEF-SIGIR2022}.




\end{abstract}




\maketitle

{
\fontsize{8pt}{8pt} 
\selectfont
\textbf{ACM Reference Format:}
\\
Yongchun Zhu, Qiang Sheng, Juan Cao, Shuokai Li, Danding Wang, and Fuzhen Zhuang. 2022. Generalizing to the Future: Mitigating Entity Bias in Fake News Detection. In  \textit{Proceedings of the 45th International ACM SIGIR Conference on Research and Development in Information Retrieval (SIGIR '22), July 11--15, 2022, Madrid, Spain.} ACM, New York, NY, USA, 6 pages. \url{https://doi.org/10.1145/3477495.3531816} }

\section{Introduction}
In recent years, more and more people acquire news from online social media where fake news has also been widely disseminated. According to Weibo 2021 annual report on fake news refutation, 66,251 fake news pieces were detected and highlighted on Weibo.\footnote{https://weibo.com/detail/4730194303126557} The wide spread of fake news on social media has threatened both individuals and society~\cite{sheng2021integrating,sheng2022zoom}. Therefore, automatic detection of fake news has been critical for promoting trust in the online news ecosystem~\cite{shu2017fake}.

\begin{table}[t]
  \centering
  \caption{Statistics of typical entities. \#news indicates the number of news pieces containing the entity in the left column. \%fake indicates the proportion of fake ones in all related news pieces. We see a significant difference of \%fake between the 2010-2017 and 2018 subset.}
  \setlength\tabcolsep{3pt}
    \begin{tabular}{lrr|rr}
    \toprule
          & \multicolumn{2}{c|}{\textbf{2010-2017}} & \multicolumn{2}{c}{\textbf{2018}} \\
    \textbf{Entity} & \textbf{\#news} & \textbf{\%fake} & \textbf{\#news} & \textbf{\%fake} \\
    \midrule
    \textbf{Beijing} & 543   & 51\% & 197   & 32\% \\
    \textbf{Hong Kong} & 212   & 73\% & 59    & 27\% \\
    \textbf{Nanjing} & 158   & 69\% & 51    & 8\% \\
    \textbf{Apple} & 66    & 62\% & 86    & 74\% \\
    \textbf{Samsung} & 54    & 65\% & 9     & 11\% \\
    \textbf{Donald Trump} & 29    & 3\% & 144    & 67\% \\
    \textbf{Jack Ma} & 28    & 57\% & 10    & 30\% \\
    \textbf{McDonald} & 24    & 54\% & 53    & 100\% \\
    \textbf{Huawei} & 21    & 0\% & 43    & 23\% \\
    \textbf{Lionel Messi} & 8     & 0\% & 95    & 89\% \\
    \bottomrule
    \end{tabular}%
  \label{tab:analysis}%
\end{table}%

In real-world scenarios, a fake news detector is generally trained on the existing news pieces and expected to detect fake news pieces in the future (i.e., ``future data'')~\cite{zhang2021mining}. In other words, the training and test data is unavoidably non-independent-and-identically-distributed (non-IID). However, most existing methods assume that the training and testing news pieces are sampled IID from a static news environment within the same period~\cite{shu2019defend,przybyla2020capturing,silva2021embracing,cheng2021causal,nan2021mdfend}, which is unrealistic. A previous experiment~\cite{zhang2021mining} has showed a large performance decrease (\textasciitilde10\%) of existing methods when changing to a more challenging temporal split from the ideal IID split.

We find that existing methods are at the risk of inadvertently capturing and even amplifying the unintended entity bias. Table~\ref{tab:analysis} lists the statistics of ten typical entities in the Weibo dataset~\cite{sheng2022zoom}. We see that the news pieces containing a certain entity have a strong correlation with the news veracity. For instance, from 2010 to 2017, 97\% of news pieces containing the entity `Donald Trump' are real. With training using such data, models would excessively depend on the existence of certain entities for prediction. However, due to the rapid changes of the news environment~\cite{sheng2022zoom}, the correlations between a certain entity and categories vary over time. In the 2018 subset, only 33\% of news pieces about `Donald Trump' are real. Therefore, if a model leans to unfairly predict news pieces containing those entities to a specific veracity label according to the biased statistical information, it might hardly generalize well when applied to the future data.

\begin{figure}[t]
	\centering
	\begin{minipage}[b]{1\linewidth}
		\centering
		\includegraphics[width=0.95\linewidth]{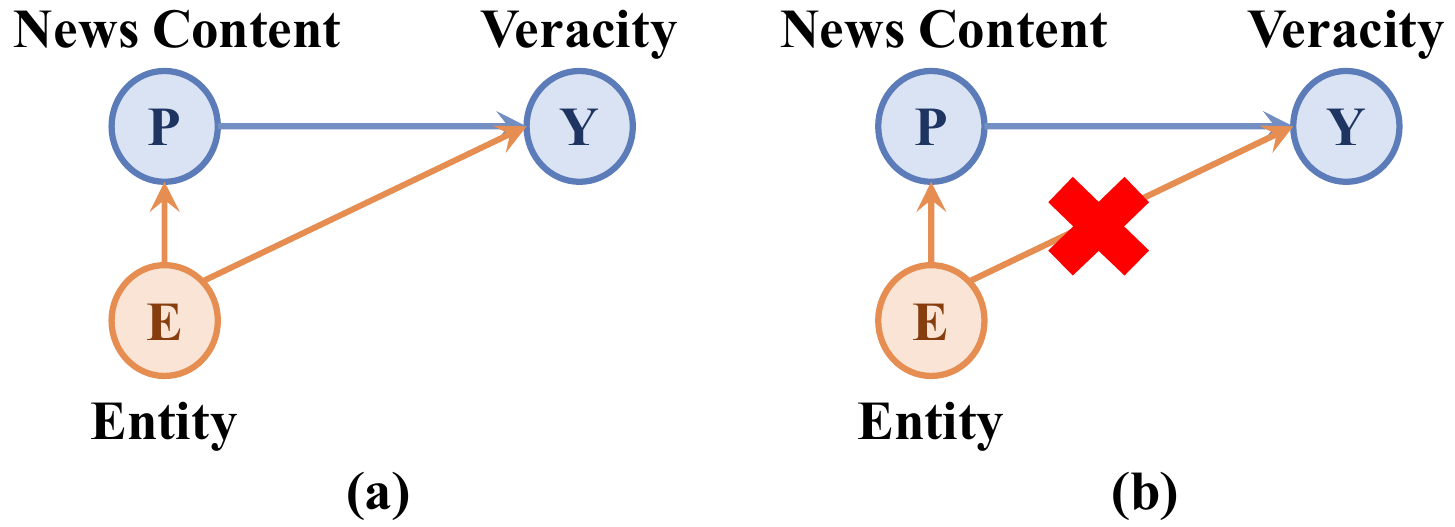}
	\end{minipage}
	\caption{(a) Causal graph for existing methods, which model effects of the news content and the confounding factor (entities). (b) Our framework aims to remove the direct effect of entities. }\label{fig:intro}
	\vspace{-0.3cm}
\end{figure}

In this paper, we mitigate the issue of entity bias from a cause-effect perspective. As shown in Figure~\ref{fig:intro}(a), existing fake news detection methods~\cite{przybyla2020capturing,zhang2021mining,sheng2021integrating} make predictions based on the overall contents of news pieces, which mix the direct effects of entities to the news veracity and the more generalizable non-entity signals such as writing style and emotions.
With the advantage of causal learning~\cite{wei2021model,qian2021counterfactual}, we propose a novel entity debiasing framework (ENDEF) which mitigates entity bias and enhance the generalization ability of base models as shown in Figure~\ref{fig:intro}(b). We specially model the direct contribution of entities to the veracity, in addition to the conventional modeling of overall contents. With explicit awareness of the entity bias during training, we remove the direct effect of related entities to perform debiased predictions. With the proposed debiasing framework, five fake news detection models in our experiment show better performance on the future data. Our contributions are as follows: 
\begin{itemize}
    \item We highlight the entity bias in fake news detection datasets and for the first time, propose to mitigate this bias for better generalization ability of fake news detectors.
    \item We design a debiasing framework that is convenient to be deployed along with different fake news detection models.
    \item We conduct both offline and online experiments to demonstrate the effectiveness of the proposed framework.
\end{itemize}

\section{Related Work}

\textbf{Fake News Detection.} It aims at classifying a news piece as real or fake. Existing methods can be roughly grouped as: content-based
and social-context-based fake news detection. Content-based methods mainly rely on news content features and factual resources to detect fake news~\cite{sheng2021integrating}, including text content~\cite{ma2019detect,nan2021mdfend,sheng2022zoom}, visual content~\cite{wang2018eann,qi2019exploiting,qi2021improving}, emotion~\cite{giachanou2019leveraging,zhang2021mining}, and evidence bases~\cite{popat2018declare,sheng2021article}. Social-context-based models exploit relevant user engagements to detect fake news~\cite{shu2017fake}, including propagation networks~\cite{nguyen2020fang,silva2021propagation2vec}, user profile~\cite{shu2018understanding,dou2021user}, and crowd feedbacks~\cite{ma2018detect,shu2019defend}. Our work falls into textual content-based methods and is closely related to \cite{wang2018eann} which extracts event-invariant features and \cite{zhang2021mining} which uses emotional signals for better generalization, but they do not explicitly highlight the bias from a temporal perspective. The experiments will show that our framework brings additional improvements even based on these methods that have considered the generalization issue.

\textbf{Model Debiasing.} The existence of the dataset bias induces models to make biased predictions, which degrades the performance on the test set~\cite{mehrabi2021survey}. Task-specific biases have been found in many areas, e.g., fact checking~\cite{schuster2019towards}, visual question-answering~\cite{agrawal2016analyzing}, recommendation~\cite{chen2020bias}. To mitigate such biases, some works perform data-level manipulations~\cite{dixon2018measuring,wei2019eda}, and others design model-level balancing mechanisms~\cite{kaneko2019gender,kang2019decoupling}. Recently, causal learning that analyzes cause-effect relationships has been utilized for model debiasing~\cite{wang2021clicks,wei2021model,qian2021counterfactual}. 
Based on our analysis on task data, we propose to mitigate the entity bias for better generalization.

\section{Methodology}
\begin{figure}[t]
	\centering
	\begin{minipage}[b]{1\linewidth}
		\centering
		\includegraphics[width=0.95\linewidth]{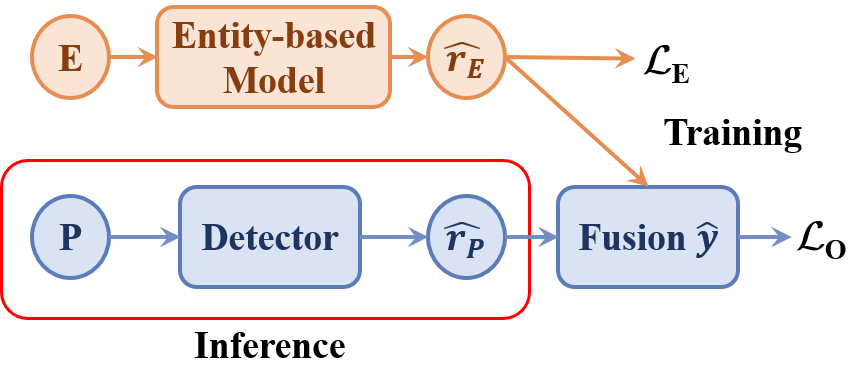}
	\end{minipage}
	\caption{The proposed entity debiasing framework (ENDEF) consists of an entity-based model and a detector. The entity-based model aims to capture the entity bias, which enables the detector to learn less biased information.}\label{fig:network}
\end{figure}

Table~\ref{fig:network} presents the proposed entity debiasing framework (ENDEF) for fake news detection, where we model the direct contribution of entities in addition to modeling overall contents, and then make debiased predictions by dropping the entity-based model.

\subsection{Problem Formulation}
Let $\mathcal{D}$ a dataset of news pieces on social media, where $P$ is a news piece in $\mathcal{D}$ containing $n$ tokens $P = \{w_1, \ldots, w_n\}$. The tokens consist of both entities and other non-entity words. The entities in the news piece $P$ are denoted as $E = \{e_1, \ldots, e_m\}$, where $m$ indicates the number of entities and each $e$ represents an entity, e.g., a person, a location. To recognize the entities, we use a public tool TexSmart~\cite{zhang2020texsmart, liu2021texsmart}\footnote{\url{https://ai.tencent.com/ailab/nlp/texsmart/en/inde.html}. We use v0.2.0 (Large).}. Each news piece has a ground-truth label $y \in \{0, 1\}$, where $1$ and $0$ denote the news piece is fake and real, respectively. Given a news piece $P$, a fake news detection detector aims to predict whether the label of $P$ is $1$ or $0$. 

\subsection{Model}
We propose a debiasing framework for fake news detection based on the causal graph in Figure~\ref{fig:intro}(b), which improves the generalization ability of base detectors by mitigating entity bias. Our framework models two cause-effect paths $E \rightarrow Y$ and $E \rightarrow P \rightarrow Y$ (Figure~\ref{fig:network}).

The distributions of entity lists $E$ may bring spurious correlations (e.g., the spurious “Lionel Messi”-to-“real class” mapping) during model training.
To explicitly model the potentially negative influence of entity bias, we train an entity-based model \textit{only} with $E$ as input to represent the direct effect from the entities $E \rightarrow Y$:
\begin{equation}
    \hat{r}_E = f_E(\{e_1, \ldots, e_m\}),
\end{equation}
where $f_E$ is a deep network. $\hat{r}_E$ is an entity-biased logit prediction.

Generally, a fake news detector takes all tokens as input, including both entities and non-entity words:
\begin{equation}
    \hat{r}_P = f_P(\{w_1, \ldots, w_n\}),
\end{equation}
where $f_P$ is a fake news detector. Note that our framework is model-agnostic, and $f_P$ can be implemented with diverse models for this task~\cite{wang2018eann,shu2019defend,zhang2021mining,nan2021mdfend}. $\hat{r}_P$ indicates its logit prediction, which represents the effect of the path $E \rightarrow P \rightarrow Y$.

The final probability prediction is aggregated with the two parts $\hat{r}_E$ and $\hat{r}_P$, formulated as:
\begin{equation}
    \hat{y} = \sigma(\alpha \hat{r}_P + (1 - \alpha) \hat{r}_E),
\end{equation}
where the $\sigma(\cdot)$ indicates the Sigmoid function and $\alpha$ is a hyper-parameter to balance the two terms. We train the overall framework with the cross-entropy loss:
\begin{equation}
    \mathcal{L}_O = \sum_{(P, y) \in \mathcal{D}} -y\log(\hat{y}) - (1 - y) \log(1 - \hat{y}).
\end{equation}

To achieve the effect of the entity module, we utilize an auxiliary loss, which applies additional supervision over the prediction of the entity-based model:
\begin{equation}
    \mathcal{L}_E = \sum_{(P, y) \in \mathcal{D}} -y\log(\sigma(\hat{r}_E)) - (1 - y) \log(1 - \sigma(\hat{r}_E)).
\end{equation}

The overall loss function can be formulated as:
\begin{equation}
    \mathcal{L} = \mathcal{L}_O + \beta \mathcal{L}_E, 
\end{equation}
where the $\beta$ denotes a hyper-parameter, which is set as 0.2 in this paper. This training procedure can make the entity-based model focus on learning the entity bias. Meanwhile, it enable the detector to learn less biased information.

This training procedure forces the entity-based model to focus on learning to detect fake news with \textit{only} the entities provided and thus fit the entity bias in the training set well. Meanwhile, it enables the fake news detector to learn less biased information by encouraging the two modules to capture different signals (entity-based and non-entity-based, respectively).

\begin{table}[t]
  \centering
  \caption{Statistics of the datasets.}
    \begin{tabular}{lrrrrrr}
    \toprule
    \multirow{2}[4]{*}{\textbf{Dataset}} & \multicolumn{3}{c}{\textbf{Weibo}} & \multicolumn{3}{c}{\textbf{GossipCop}} \\
\cmidrule{2-7}          & Train & Val   & Test  & Train & Val   & Test \\
    \midrule
    \#Fake  & 2,561 & 499   & 754   & 2,024 & 604   & 601 \\
    \#Real  & 7,660 & 1,918 & 2,957 & 5,039 & 1,774 & 1,758 \\
    Total & 10,221 & 2,417 & 3,711 & 7,063 & 2,378 & 2,359 \\
    \bottomrule
    \end{tabular}%
  \label{tab:dataset}%
\end{table}%

\subsection{Inference}
To mitigate entity bias for better generalization ability, the key is to remove the direct effect via path $E \rightarrow Y$ from the prediction $\hat{y}$. Since $\hat{y}$ can be seen as the total effect and $\hat{r}_E$ is the natural direct effect of entities to the news veracity label, the remaining $\hat{r}_P$ is actually a prediction based on less biased information. Therefore, we could mitigate the entity bias by simply using $\sigma(\hat{r}_P)$ during inference. Note that the detector in Figure~\ref{fig:network} is not limited to specific models, making our debiasing framework be compatible with diverse base models to improve their generalization ability.

\subsection{Data Augmentation}
Data augmentation technique has shown its power for alleviating overfitting~\cite{li2022learning}. To further improve the generalization ability of the models, we adopt two types of token-level augmentation techniques, including drop (deleting the selected token) and mask (replacing the selected token with a special token \texttt{[MASK]}). In addition, we apply two augmentation policies: (1) randomly drop or mask words with the probability of $p$; and (2) randomly drop or mask entities with the probability of $p$. In the training stage, we randomly adopt one augmentation policy for each sample.
\section{Experiments}
We experimentally answer the following research questions:
\begin{itemize}
    \item[\textbf{RQ1}] Can our framework improve the generalization ability of fake news detection on future data?
    \item[\textbf{RQ2}] Can this debiasing framework bring improvement to the performance of the real-world online system?
    \item[\textbf{RQ3}] How does the mitigation of entity bias improve the performance of base models? 
\end{itemize}

\begin{table*}[t]
  \centering
  \setlength\tabcolsep{4pt}
  \caption{Offline performance comparison of base models with and without the debiasing framework. The better result in each group using the same base model are in boldface. The marker * indicates that the improvement is statistically significant compared with the best baseline (paired t-test with p-value < 0.05).}
    \begin{tabular}{lcccccccccccc}
    \toprule
    \multirow{2}[1]{*}{\textbf{Method}} & \multicolumn{6}{c}{\textbf{Weibo}}         & \multicolumn{6}{c}{\textbf{GossipCop}} \\
    \cmidrule(lr){2-7} \cmidrule(lr){8-13}  
    & macF1    & Acc   & AUC   & spAUC & F1$_{\text{real}}$ & F1$_{\text{fake}}$ & macF1    & Acc   & AUC   & spAUC & F1$_{\text{real}}$ & F1$_{\text{fake}}$ \\
    \midrule
    BiGRU & 0.7172  & 0.8214  & 0.8354  & 0.6636  & 0.8887  & 0.5456  & 0.7730  & 0.8379  & 0.8634  & 0.7358  & 0.8943  & 0.6516  \\
    \ \ \ \ w/ ENDEF & \textbf{0.7318*} & \textbf{0.8286*} & \textbf{0.8446*} & \textbf{0.6802*} & \textbf{0.8929*} & \textbf{0.5707*} & \textbf{0.7842*} & \textbf{0.8465*} & \textbf{0.8669} & \textbf{0.7472*} & \textbf{0.8989*} & \textbf{0.6696*} \\
    EANN  & 0.7162  & 0.8197  & 0.8276  & 0.6649  & 0.8875  & 0.5448  & 0.7926  & 0.8517  & 0.8765  & 0.7586  & 0.9033  & 0.6820  \\
    \ \ \ \ w/ ENDEF & \textbf{0.7370*} & \textbf{0.8316*} & \textbf{0.8398*} & \textbf{0.6886*} & \textbf{0.8947*} & \textbf{0.5793*} & \textbf{0.7937} & \textbf{0.8526} & \textbf{0.8836*} & \textbf{0.7620*} & \textbf{0.9039} & \textbf{0.6835} \\
    BERT & 0.7601  & 0.8474  & 0.8754  & 0.7102  & 0.9048  & 0.6155  & 0.7873  & 0.8439  & 0.8781  & 0.7579  & 0.8968  & 0.6778  \\
   \ \ \ \  w/ ENDEF & \textbf{0.7714*} & \textbf{0.8550*} & \textbf{0.8824*} & \textbf{0.7257*} & \textbf{0.9096*} & \textbf{0.6332*} & \textbf{0.7969*} & \textbf{0.8496*} & \textbf{0.8853*} & \textbf{0.7663*} & \textbf{0.8994} & \textbf{0.6944*} \\
    MDFEND & 0.7051  & 0.7786  & 0.8301  & 0.6691  & 0.8519  & 0.5584  & 0.7905  & \textbf{0.8518}  & 0.8712  & 0.7543  & \textbf{0.9037}  & 0.6772  \\
    \ \ \ \ w/ ENDEF & \textbf{0.7313*} & \textbf{0.8057*} & \textbf{0.8490*} & \textbf{0.6879*} & \textbf{0.8724*} & \textbf{0.5902*} & \textbf{0.7970*} & 0.8517 & \textbf{0.8824*} & \textbf{0.7627*} & 0.9023 & \textbf{0.6916*} \\
    BERT-Emo & 0.7586  & 0.8438  & 0.8743  & 0.7061  & 0.9019  & 0.6154  & 0.7912  & 0.8455  & 0.8800  & 0.7631  & 0.8974  & 0.6849  \\
    \ \ \ \ w/ ENDEF & \textbf{0.7731*} & \textbf{0.8584*} & \textbf{0.8838*} & \textbf{0.7278*} & \textbf{0.9121*} & \textbf{0.6341*} & \textbf{0.8010*} & \textbf{0.8520*} & \textbf{0.8855*} & \textbf{0.7674*} & \textbf{0.9020*} & \textbf{0.6987*} \\
    \bottomrule
    \end{tabular}%
  \label{tab:offline}%
\end{table*}%

\begin{table}[htbp]
  \centering
  \setlength\tabcolsep{3pt}
   \caption{Results on the online data. Each row indicates the relative improvement with our ENDEF framework over the baselines.}
    \begin{tabular}{ccccccc}
    \toprule
    Method & macF1 & Acc   & AUC   & spAUC & F1$_{\text{real}}$  & F1$_{\text{fake}}$ \\
    \midrule
    BiGRU & 2.56\% & 2.02\% & 4.02\% & 3.26\% & 1.12\% & 6.45\% \\
    EANN & 0.57\% & -0.77\% & 1.64\% & 1.02\% & -0.44\% & 3.26\% \\
    BERT & 0.60\% & 0.57\% & 0.44\% & 0.32\% & 0.33\% & 1.12\% \\
    MDFEND & 2.57\% & 2.02\% & 1.14\% & 0.95\% & 1.19\% & 5.50\% \\
    BERT-Emo & 0.68\% & 0.78\% & 0.22\% & 0.78\% & 0.13\% & 2.44\% \\
    \bottomrule
    \end{tabular}%
  \label{tab:online}%
\end{table}%

\subsection{Experimental Settings}
\textbf{Datasets.} A Chinese dataset and an English dataset are adopted for evaluation. For the Chinese dataset, we adopt the Weibo dataset~\cite{sheng2022zoom} from 2010 to 2018.\footnote{https://github.com/ICTMCG/News-Environment-Perception/}
For the English dataset, we adopt the GossipCop data of FakeNewsNet~\cite{shu2020fakenewsnet}.\footnote{https://github.com/KaiDMML/FakeNewsNet} 
To simulate the real-world temporal scenarios, we adopt the \textit{temporal} split strategy. The most recent 40\% news pieces are randomly included in the test and validation sets. The remaining 60\% of news pieces serve as the training set. Specifically, for the Weibo dataset, the time period of the training set is from 2010 to 2017, and the samples in the test and validation sets are posted in 2018. For the GossipCop dataset, the time period of the training set is from 2000 to 2017, and all samples in the test and validation sets are posted in 2018. Table~\ref{tab:dataset} shows the statistics of the datasets.

\textbf{Base Models.} Technically, our framework is model-agnostic, which could coordinate with various fake news detectors. Here we select five representative content-based detection methods as our base models. 
\begin{itemize}
    \item BiGRU~\cite{cho2014properties} is widely used in many existing works of our task for text encoding~\cite{nan2021mdfend,zhang2021mining}. We implement a one-layer BiGRU with a hidden size of 768. Then, we utilize a mask attention layer to aggregate all the hidden states as representations of posts which are further fed into an MLP for prediction.
    \item EANN~\cite{wang2018eann} is a model that tries to distract the fake news detection model from memorizing event-specific features. It uses TextCNN for text representation and adds an auxiliary task of event classification for adversarial learning using gradient reversal layer. The complete EANN is a multi-modal model but we here use its text-only version. For TextCNN, the window sizes are \{1, 2, 3, 5, 10\}. The labels for the auxiliary event classification task are derived by clustering according to the publication year.
    \item BERT~\cite{kenton2019bert, cui2019pre} is a popular pre-training model. We utilize BERT to encode tokens of news content and feed the extracted average embedding into an MLP to obtain the final prediction. For the GossipCop dataset, we adopt the original BERT model~\cite{kenton2019bert}. For the Weibo dataset, we adopted a modified BERT model~\cite{cui2019pre}.
    \item MDFEND~\cite{nan2021mdfend} is the latest multi-domain text-based fake new detection model which utilizes a Domain Gate to select useful experts of MoE. We adopt the same TextCNN structure for all experts. The number of experts is set as 5. In this paper, we utilize the publication year as the domain label.
    \item BERT-Emo~\cite{zhang2021mining} combines emotional features and the BERT detector for fake news detection. As we focus on the contents rather than social contexts, we adopt a simplified version where emotions in comments are not considered.
\end{itemize}

\textbf{Evaluation Metrics.} Following most existing works~\cite{wang2018eann,shu2019defend,zhang2021mining}, we report Area Under ROC (AUC), accuracy (Acc), macro F1 score (macF1) and the F1 scores of fake and real class (F1$_{\text{fake}}$ and F1$_{\text{real}}$). In addition, as the datasets are skewed (real: fake $\approx$ 3: 1), a fake news detector should detect fake news without misclassifying real news as possible. Formally speaking, we should improve the true positive rate (TPR) on the basis of low false positive rate (FPR). Therefore, following~\cite{mcclish1989analyzing,zhu2020modeling,sheng2022zoom}, we bring a metric into the evaluation of fake news detectors named standardized partial AUC ($\mathrm{spAUC_{FPR \leq maxfpr}}$):
\begin{equation}
\begin{split}
    \mathrm{spAUC_{FPR \leq maxfpr}} &= \frac{1}{2} \left (1 + \frac{ \mathrm{AUC_{FPR \leq maxfpr} } - \mathrm{minarea}}{\mathrm{maxarea} - \mathrm{minarea}}\right ), \\
    \mathrm{where} \quad \mathrm{maxarea} &= \mathrm{maxfpr},\\
    \mathrm{minarea} &= \frac{1}{2} \times \mathrm{maxfpr}^2.
\end{split}
\end{equation}
In practice, we require FPR to be less than 10\%. Hence, in this paper, we use $\mathrm{spAUC_{FPR \leq 10\%}}$ for all experiments.

\textbf{Implementation Details.} \textls[10]{We did not perform any dataset-specific} tuning except early stopping on validation sets. For all methods, the initial learning rate for the Adam optimizer~\cite{kingma2014adam} was tuned by grid search in [1e-6, 1e-2]. We performed a grid search in [0,1] with the step of 0.1 for searching the best hyper-parameter $\alpha$ and we found 0.8 was the best value. The mini-batch size is 64. For the MLPs in these methods, we employed the ReLU function and set the dimension of the hidden layer as 384. The maximum sequence lengths of GossipCop and Weibo datasets were set as 170. We set the probability of applying data augmentation $p$ as 0.1. We ran all methods on both two datasets for ten times and report the average scores for all metrics.

\begin{figure}[t]
	\centering
	\begin{minipage}[b]{1\linewidth}
		\centering
		\includegraphics[width=0.95\linewidth]{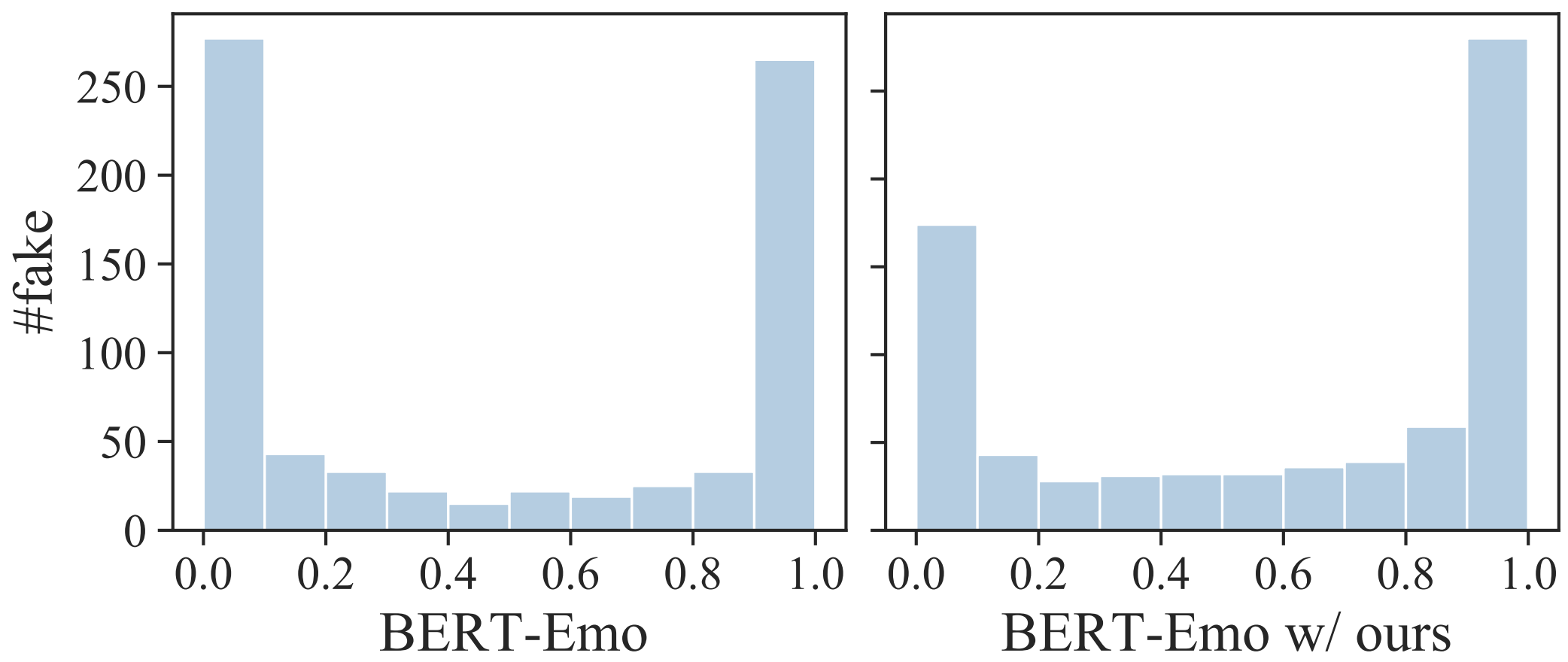}
	\end{minipage}
	\caption{The distribution of prediction scores on fake news pieces in the Weibo test set. The average prediction scores of BERT-Emo and BERT-Emo w/ ours are 0.49 and 0.58, respectively.}\label{fig:analysis}
\end{figure}

\subsection{Results}
\subsubsection{Offline Experiments (RQ1).} We conduct offline experiments with two datasets. Specifically, we apply the proposed framework upon various base models. The performance of these base models with and without the framework is shown in Table~\ref{tab:offline}. From these results, we have the following observations: 
\begin{itemize}
    \item With the help of the proposed framework, most base models show a significant improvement in most metrics, which demonstrates the effectiveness of the proposed framework on future data. In addition, it also testifies ENDEF is a general framework which can be applied upon various base models.
    \item For most metrics, the performance improvement in the Weibo dataset is larger than that in the GossipCop dataset. We attribute such a difference between the two datasets to the length of a news piece. The average length of news pieces in the Weibo dataset is 120, while the average length of the GossipCop dataset is 606. The longer news piece would have more informative patterns, e.g., writing style, emotion, which alleviating the influence of entities.
\end{itemize}

\subsubsection{Online Experiments (RQ2).} We tested these models on a dump of ten-month data in 2021 from our Chinese fake news detection system. Different from the offline datasets, this online data set is much more skewed (30,977 real: 774 fake $\approx$ 40:1). Due to the restriction of business rules, we cannot report the absolute performance of these methods. Instead, we report the relative improvement of our proposed framework compared with different base models in Table~\ref{tab:online}, and each row indicates the relative improvement of the base model with the proposed framework over the base model. 
The online results show the proposed debiasing framework can improve the base models in a highly skewed scenario, which demonstrates the importance of alleviating the influence of entity bias in real-world detection systems. The best-performing model \textit{BERT-Emo w/ ours} has been deployed in our online system which handles thousands of suspicious news pieces every day.

\begin{figure}[t]
	\centering
	\begin{minipage}[b]{1\linewidth}
		\centering
		\includegraphics[width=0.95\linewidth]{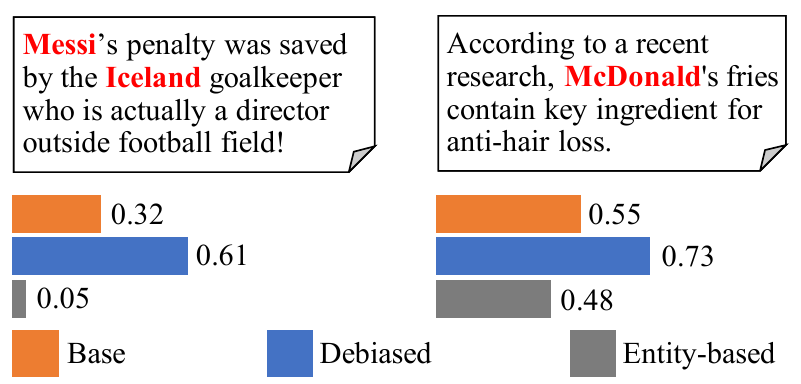}
	\end{minipage}
	\caption{Two fake news cases. The entities are {\color{red}{\textbf{boldfaced}}}. The lengths of bars represent the probability predictions of the base model (BERT-Emo), our debiased model, and the entity-based model. Our debiased model shows higher confidence in predicting the two samples as fake.}\label{fig:case}
\end{figure}

\subsubsection{Analysis (RQ3).} Figure~\ref{fig:analysis} presents the distribution of prediction scores of BERT-Emo and BERT-Emo w/ ours on fake news in the Weibo test set, which demonstrates the base model with our framework can detect fake news more accurately (higher average scores of fake news). In addition, we show two fake news cases in Figure~\ref{fig:case}. \textit{Debiased} indicates the final debiased prediction $\sigma(\hat{r}_P)$, and \textit{entity-based} represents the prediction of the entity-based model.
The two cases show that the entity-based model can capture the biased information, which enables the detector trained with our framework to make debiased predictions.

\section{Conclusion and Future Work}
We proposed a novel entity debiasing framework (ENDEF) which improves the generalization ability of base fake news detectors on future data by mitigating the largely overlooked entity bias in existing works. Specifically, we designed a debiasing framework based on a causal graph of entities, news contents, and news veracity, where the direct effect of entities to the news veracity is explicitly modeled during training the base fake news detector and finally removed during inference to perform debiased prediction. Both offline and online experiments demonstrated the effectiveness of our proposed framework.

To the best of our knowledge, this is the first work to explicitly focus on improving the generalization ability of fake news detection models to the future data, which is a practical problem in the real-world detection system. We believe further exploration is required for a deeper understanding of the gap between news pieces in different periods.  In the future, we plan to explore: (1) adapting unbiased model to the future news environment; (2) investigating the common features between news pieces of different periods; (3)  extending the proposed framework ENDEF from the text-only detection to multi-modal and social graph-based detection.

\begin{acks}
The authors thank the anonymous reviewers for their insightful and valuable comments. The research work is supported by the National Key Research and Development Program of China (2021AAA0140203), the Zhejiang Provincial Key Research and Development Program of China (2021C01164), and the National Natural Science Foundation of China (62176014).
\end{acks}

\balance
\bibliographystyle{ACM-Reference-Format}
\bibliography{sample-base}


\end{document}